\documentclass[sigconf]{acmart}
\usepackage{booktabs} 
\makeatletter                   %
\def\mdseries@tt{m}             %
\makeatother                    %

\usepackage[plain]{fancyref}
\usepackage[draft=true]{minted} %
\usepackage{color}
\usepackage{breakurl}           %

 \usepackage{lipsum}
\usepackage[final]{pdfpages}
\usepackage{setspace}
\usepackage{epsf}
\usepackage{epsfig}
\usepackage{graphicx}
\usepackage{algorithmic}
\usepackage{amsmath}
\usepackage{letltxmacro}

\usepackage{enumitem}
\usepackage{tikz}
\usepackage{libpaper/tightenum}

\usepackage{url}
\usepackage{multirow}

  \renewenvironment{thebibliography}[1]{%
    \begin{oldthebibliography}{#1}%
      \setlength{\parskip}{0ex}%
      \setlength{\itemsep}{0ex}%
  }%
  {%
    \end{oldthebibliography}%
  }

\newwrite\arxivdeps
\immediate\openout\arxivdeps=\jobname-arxivdeps.log

\newcommand\verifymarkedforarxivfile[1]{%
\ifdefined\arxivbuild
\else
\IfFileExists{#1}%
{}%
{\GenericWarning{Marked file (#1) for inclusion in arxiv build doesn't exist}}%
\fi%
}

\newcommand\markforarxiv[1]{%
\verifymarkedforarxivfile{#1}%
\write\arxivdeps{IncludeInArxiv: #1}%
}

\DeclareUrlCommand\UScore{\urlstyle{rm}}

\LetLtxMacro\oldincludegraphics\includegraphics
\renewcommand{\includegraphics}[2][]{%
\markforarxiv{#2}%
\oldincludegraphics[#1]{#2}}

\LetLtxMacro\oldincludepdf\includepdf
\renewcommand{\includepdf}[2][]{%
\markforarxiv{#2}%
\oldincludepdf[#1]{#2}}

\def\cfigure[#1,#2,#3]{
\begin{figure}
\vspace*{0mm}
\begin{center}

\includegraphics[width=3in]{#1} 
\vspace*{-3mm}\caption[]{#2
} \label{#3}
 
\vspace*{-5mm}
\end{center}
\end{figure}}

\def\cfigurefour[#1,#2,#3]{
\begin{figure}
\vspace*{0mm}
\begin{center}

\includegraphics[width=4in]{#1} 
\vspace*{-3mm}\caption[]{#2
} \label{#3}
 
\vspace*{-5mm}
\end{center}
\end{figure}}

\def\cfiguretemp[#1,#2,#3]{
\begin{figure}
\vspace*{0mm}
\begin{center}

\includegraphics[width=3.5in]{#1} 
\vspace*{-3mm}\caption[]{#2
} \label{#3}
 
\vspace*{-5mm}
\end{center}
\vspace*{-2mm}
\end{figure}}

\def\wfigure[#1,#2,#3]{
\begin{figure*}
\vspace*{0mm}
\begin{center}
 \includegraphics[width=\textwidth]{#1} 
 \vspace*{-3mm}\caption[]{#2
} \label{#3}
 
\end{center}
\end{figure*}}

\def\threefigure[#1,#2,#3,#4,#5]{
\begin{figure*}
\vspace*{0mm}
\begin{center}

\begin{tabular}{ccc}
\includegraphics[width=2in]{#1} & \includegraphics[width=2in]{#2} &  \includegraphics[width=2in]{#3} \\
(a) & (b) & (c) \\
\end{tabular}
\vspace*{-3mm}\caption[]{#4
} \label{#5}

\vspace*{-5mm}
\end{center}
\vspace*{-2mm}
\end{figure*}}

\def\dcfigure[#1,#2,#3,#4,#5,#6]{
{
\begin{figure*}
\begin{center}
\begin{minipage}[c]{\columnwidth}{
\includegraphics[width=\columnwidth]{#1} 
\vspace*{0mm}\caption[]{#2} \label{#3} \
}\end{minipage}\hspace*{\columnsep}\
\begin{minipage}[c]{\columnwidth}{
\includegraphics[width=\columnwidth]{#4} 
\vspace*{0mm}\caption[]{#5}\label{#6} \
}\end{minipage}
\end{center}
\end{figure*}
}
}

\def\scfigure[#1,#2,#3]{
{
\begin{figure*}
\begin{center}
\begin{minipage}[c]{3.5in}{
\includegraphics[width=3.5in]{#1} 
}\end{minipage}
\caption[]{#2} \label{#3} \
\end{center}
\end{figure*}
}
}

\def\tableByTable[#1,#2,#3,#4,#5,#6]{
{
\begin{table*}
\begin{center}
\begin{minipage}[c]{3in}{
\centering
{#1}
\vspace*{0mm}\tabcaption[]{#2}\label{#3} \
}\end{minipage}\hspace*{\columnsep}\
\begin{minipage}[c]{3in}{
\centering
{#4}
\vspace*{0mm}\tabcaption[]{#5}\label{#6} \
}\end{minipage}
\end{center}
\end{table*}
}
}

\def\figureByTable[#1,#2,#3,#4,#5,#6]{
{
\begin{figure*}
\begin{center}
\begin{minipage}[c]{3in}{
\centering
\includegraphics[width=\textwidth]{#1}
\vspace*{0mm}\figcaption[]{#2} \label{#3} \
}\end{minipage}\hspace*{\columnsep}\
\begin{minipage}[c]{3.3in}{
\centering
{#4}
\vspace*{0mm}\tabcaption[]{#5}\label{#6} \
}\end{minipage}
\end{center}
\end{figure*}
}
}

\def\tableByFigure[#1,#2,#3,#4,#5,#6]{
{
\begin{figure*}
\begin{center}
\begin{minipage}[c]{4.3in}{
\centering
{#1}
\vspace*{0mm}\tabcaption[]{#2} \label{#3} \
}\end{minipage}\hspace*{\columnsep}\
\begin{minipage}[c]{2.2in}{
\centering
\includegraphics[width=\textwidth]{#4}
\vspace*{-0.35in}\caption[]{#5}\label{#6} \
}\end{minipage}
\end{center}
\end{figure*}
}
}

\def\doublecfigure[#1,#2,#3,#4]{
{
\begin{figure}
\begin{center}
\begin{minipage}[c]{1.5in}{
\begin{center}
\includegraphics[width=1.5in]{#1}%
\end{center}
}\end{minipage}\hspace*{1em}\
\begin{minipage}[c]{1.5in}{
\begin{center}
\includegraphics[width=1.5in]{#2}%
\end{center}
}\end{minipage}
\vspace*{0mm}\caption[]{#3} \label{#4} \
\end{center}
\end{figure}
}
}

\def\qcfigure[#1,#2,#3,#4,#5,#6]{
{
\begin{figure*}
\vspace*{0.2in}\
\begin{center}
\begin{minipage}[c]{3in}{
\includegraphics[width=3in]{#1} 
\vspace*{-3mm}
}
\end{minipage}\hspace*{0.5in}\
\begin{minipage}[c]{3in}{
\includegraphics[width=3in]{#2} 
\vspace*{-3mm}
}\end{minipage}

\begin{minipage}[c]{3in}{
\includegraphics[width=3in]{#3} 
\vspace*{-3mm}
}
\end{minipage}\hspace*{0.5in}\
\begin{minipage}[c]{3in}{
\includegraphics[width=3in]{#4} 
\vspace*{-3mm}
}\end{minipage}
\end{center}
\caption[]{#5}\label{#6}
\end{figure*}
}
}

\def\twfigure[#1,#2,#3,#4,#5]{
{
\begin{figure*}
\vspace*{0.2in}\
\begin{center}
\begin{minipage}[c]{6.5in}{
\includegraphics[width=6.5in]{#1} 
\vspace*{-3mm}
}
\end{minipage}

\begin{minipage}[c]{6.5in}{
\includegraphics[width=6.5in]{#2} 
\vspace*{-3mm}
}\end{minipage}

\begin{minipage}[c]{6.5in}{
\includegraphics[width=6.5in]{#3} 
\vspace*{-3mm}
}
\end{minipage}
\end{center}
\caption[]{#4}\label{#5}
\end{figure*}
}
}

\def\dwfigure[#1,#2,#3,#4]{
{
\begin{figure*}
\vspace*{0.2in}\
\begin{center}
\begin{minipage}[c]{6.5in}{
\includegraphics[width=6.5in]{#1} 
\vspace*{-3mm}
}
\end{minipage}

\begin{minipage}[c]{6.5in}{
\includegraphics[width=6.5in]{#2} 
\vspace*{-3mm}
}\end{minipage}

\end{center}
\caption[]{#3}\label{#4}
\end{figure*}
}
}

\def\dssfigure[#1,#2,#3,#4,#5,#6]{
{
\begin{figure*}
\vspace*{0.2in}\
\begin{center}
\begin{minipage}[c]{4in}{
\includegraphics[width=4in]{#1}
\vspace*{-3mm}\caption[]{#2} \label{#3} \
}\end{minipage}\hspace*{0.5in}\
\begin{minipage}[c]{2in}{
\includegraphics[width=2in]{#4}
\vspace*{-3mm}\caption[]{#5}\label{#6} \
}\end{minipage}
\end{center}
\vspace*{-0.4in}\
\end{figure*}
}
}

\def\dsfigure[#1,#2,#3,#4,#5,#6]{
{
\begin{figure*}
\vspace*{0.2in}\
\begin{center}
\begin{minipage}[c]{3in}{
\includegraphics[width=3in]{#1}
\vspace*{-3mm}\caption[]{#2} \label{#3} \
}\end{minipage}\hspace*{0.5in}\
\begin{minipage}[c]{3in}{
\hspace*{0.5in}\
\includegraphics[height=3in]{#4}
\vspace*{-3mm}\caption[]{#5}\label{#6} \
}\end{minipage}
\end{center}
\vspace*{-0.4in}\
\end{figure*}
}
}

\def\dsyfigure[#1,#2,#3,#4,#5,#6]{
{
\begin{figure*}
\vspace*{0.2in}\
\begin{center}
\begin{minipage}[c]{2.5in}{
\includegraphics[height=2.5in]{#1}
\vspace*{-3mm}\caption[]{#2} \label{#3} \
}\end{minipage}\hspace*{0.5in}\
\begin{minipage}[c]{2.5in}{
\includegraphics[height=2.5in]{#4}
\vspace*{-3mm}\caption[]{#5}\label{#6} \
}\end{minipage}
\end{center}
\vspace*{-0.4in}\
\end{figure*}
}
}

\def\dyfigure[#1,#2,#3,#4,#5,#6]{
{
\begin{figure*}
\vspace*{0.2in}\
\begin{center}
\begin{minipage}[c]{3in}{
\includegraphics[height=3in]{#1} 
\vspace*{-3mm}\caption[]{#2} \label{#3} \
}\end{minipage}\hspace*{0.5in}\
\begin{minipage}[c]{3in}{
\includegraphics[height=3in]{#4} 
\vspace*{-3mm}\caption[]{#5}\label{#6} \
}\end{minipage}
\end{center}
\vspace*{-0.4in}\
\end{figure*}
}
}

\def\dyoldfigure[#1,#2,#3,#4,#5,#6]{
{
\begin{figure*}
\vspace*{0.2in}\
\begin{center}
\begin{minipage}[c]{3in}{
\epsfysize=2.0in\
\hspace{0.5in}\
\epsfbox{#1}
\vspace*{-3mm}\caption[]{#2} \label{#3} \
}\end{minipage}\hspace*{0.25in}\
\begin{minipage}[c]{3in}{
\epsfysize=2.0in\
\hspace{0.5in}\
\epsfbox{#4}
\vspace*{-3mm}\caption[]{#5}\label{#6} \
}\end{minipage}
\end{center}
\vspace*{-0.4in}\
\end{figure*}
}
}

\def\cfiguredouble[#1,#2,#3,#4]{
\begin{figure}
\vspace*{0.2in}\
\begin{center}
\begin{minipage}[c]{1.5in}{
\epsfxsize=1.5in\
\epsfbox{#1}
}\end{minipage}\hspace*{0.1in}\
\begin{minipage}[c]{1.5in}{
\epsfxsize=1.5in\
\vspace{0.1in}\epsfbox{#2}
}\end{minipage}\vspace*{-0.10in} \caption[]{#3}\label{#4}
\end{center}
\vspace*{-0.4in}\
\end{figure}
}

\def\wpfigure[#1,#2,#3,#4]{
\begin{figure*}
\vspace*{4mm}
\begin{center}

\includegraphics[width=#4]{#1} 

\vspace*{-3mm}\caption[]{#2
} \label{#3}

\vspace*{-5mm}
\end{center}
\end{figure*}}

\def\wprfigure[#1,#2,#3,#4,#5]{
\begin{figure*}
\vspace*{4mm}
\begin{center}

\includegraphics[width=#4, angle=#5]{#1} 

\vspace*{-3mm}\caption[]{#2
} \label{#3}

\vspace*{-5mm}
\end{center}
\end{figure*}}

\def\DoubleFigureWSlide[#1,#2,#3,#4,#5,#6,#7,#8,#9]{
\begin{figure*}
\vspace*{#9}
\begin{center}
\begin{minipage}{#4}
\includegraphics[width=#4]{#1}
\vspace*{-3mm}\caption{#2
}\label{#3}
\end{minipage}
\hspace{2em}
\begin{minipage}{#8}
\includegraphics[width=#8]{#5}
\vspace*{-3mm}\caption{#6
}\label{#7}
\end{minipage}
\vspace*{-5mm}
\end{center}
\end{figure*}
}

\def\DoubleFigureW[#1,#2,#3,#4,#5,#6,#7,#8]{
\begin{figure*}
\vspace*{0in}
\begin{center}
\begin{minipage}{#4}
\includegraphics[width=#4]{#1}
\vspace*{-3mm}\caption{#2
}\label{#3}
\end{minipage}
\hspace{2em}
\begin{minipage}{#8}
\includegraphics[width=#8]{#5}
\vspace*{-3mm}\caption{#6
}\label{#7}
\end{minipage}
\vspace*{-5mm}
\end{center}
\end{figure*}
}

\def\DoubleFigureWHack[#1,#2,#3,#4,#5,#6,#7,#8]{
\begin{figure*}
\vspace*{0in}
\begin{center}
\begin{minipage}{3in}
\includegraphics[width=#4]{#1}
\vspace*{-3mm}\caption{#2
}\label{#3}
\end{minipage}
\hspace{2em}
\begin{minipage}{3in}
\includegraphics[width=#8]{#5}
\vspace*{-3mm}\caption{#6
}\label{#7}
\end{minipage}
\vspace*{-5mm}
\end{center}
\end{figure*}
}

\def\ddcfigure[#1,#2,#3,#4]{
\begin{figure*}
\vspace*{0.2in}\
\begin{center}
\begin{minipage}[c]{\columnwidth}{
\includegraphics[width=\columnwidth]{#1} 
}\end{minipage}\hspace{0.5in}\
\begin{minipage}[c]{\columnwidth}{
\includegraphics[width=\columnwidth]{#2} 
}\end{minipage} \caption[]{#3}\label{#4}
\end{center}
\end{figure*}
}

\def\ddcfigureSlide[#1,#2,#3,#4,#5]{
\begin{figure*}
\vspace*{#5}\
\begin{center}
\begin{minipage}[c]{3in}{
\includegraphics[height=3in]{#1} 
}\end{minipage}\hspace{0.5in}\
\begin{minipage}[c]{3in}{
\includegraphics[height=3in]{#2} 
}\end{minipage}\vspace*{-0.10in} \caption[]{#3}\label{#4}
\end{center}
\vspace*{-0.4in}\
\end{figure*}
}

\def\cxfigure[#1,#2,#3]{
\begin{figure}
\vspace*{4mm}
\begin{center}
 
\epsfxsize=2.5in\
\epsfbox{#1}\
 
\vspace*{-0.10in}\caption[]{#2
} \label{#3}
 
\vspace*{-5mm}
\end{center}
\vspace*{-2mm}
\end{figure}}

\newif\ifremark
\long\def\remark#1{
\ifremark%
        \begingroup%
        \dimen0=\columnwidth
        \advance\dimen0 by -1in%
        \setbox0=\hbox{\parbox[b]{\dimen0}{\protect\em #1}}
        \dimen1=\ht0\advance\dimen1 by 2pt%
        \dimen2=\dp0\advance\dimen2 by 2pt%
        \vskip 0.25pt%
        \hbox to \columnwidth{%
                \vrule height\dimen1 width 3pt depth\dimen2%
                \hss\copy0\hss%
                \vrule height\dimen1 width 3pt depth\dimen2%
        }%
        \endgroup%
\fi}

\usepackage{xcolor}

\definecolor{cyanish}{rgb}{0,0.8,1.0}
\definecolor{orange}{rgb}{1.0,0.5,0.0}
\definecolor{pink}{rgb}{1.0,0.47,0.6}
\definecolor{light-gray}{gray}{0.95}
\definecolor{jiancolor}{RGB}{0,153,153}
\definecolor{mygreen}{RGB}{50,200,50}
\definecolor{pink}{rgb}{1.0,0.47,0.6}

\newcommand{\boldparagraph}[1]{\vspace*{1ex}\noindent\textbf{#1}\hspace{1em}}

\newcommand{\ignore}[1]{}

\newcommand{\myitem}[1]{\item \textbf{#1}}

\newcommand{\reffig}[1]{Figure~\ref{#1}}

\newcommand{\refsec}[1]{Section~\ref{#1}}

\newcommand{\x}[1]{$\times$}
\newcommand{\figtitle}[1]{\textbf{#1}}

\newcommand{\q}{quadcopter}

\newcommand{\qs}{quadcopters}

\renewcommand{\~}{$\sim$}

\newcommand{\lab}[1]{Lab~#1}
\newcommand{\eaglint}{QuadLint}

\renewcommand{\figtitle}[1]{#1 -- }

\renewcommand{\myitem}[1]{\item \textbf{#1:}~}

\remarktrue

\copyrightyear{2019} 
\acmYear{2019} 
\setcopyright{rightsretained} 
\acmConference[SIGCSE '19]{Proceedings of the 50th ACM Technical Symposium on Computer Science Education}{February 27-March 2, 2019}{Minneapolis, MN, USA}
\acmBooktitle{Proceedings of the 50th ACM Technical Symposium on Computer Science Education (SIGCSE '19), February 27-March 2, 2019, Minneapolis, MN, USA}
\acmDOI{10.1145/3287324.3287451}
\acmISBN{978-1-4503-5890-3/19/02}
\begin{document}

\title[Trial by Flyer: Quadcopters in 10 Weeks]{Trial by Flyer: Building Quadcopters From Scratch in a Ten-Week Capstone Course}
\author{Steven Swanson}
\orcid{0000-0002-5896-1037}
\affiliation{%
  \institution{University of Califonia, San Diego}
  \streetaddress{9500 Gilman Dr}
  \city{San Diego}
  \state{California}
  \postcode{92093-0404}
}
\email{swanson@cs.ucsd.edu}



\begin{abstract}
We describe our experience teaching an intensive capstone course in which pairs of students build the hardware and software for a remote-controller quad-rotor aircraft (i.e., a \q{} or ``drone'') from scratch in one 10-week quarter.  The course covers printed circuit board (PCB) design and assembly, basic control theory and sensor fusion, and embedded systems programming.  To reduce the workload on course staff and provide higher-quality feedback on student designs, we have implemented an automated PCB design checking tool/autograder.  We describe the course content in detail, identify the challenges it presents to students and course staff, and propose changes to further increase student success and improve the scalability of the course.

\end{abstract}

\maketitle

\section{Introduction}
\label{sec:introduction}

Compute-capable devices and robots are permeating every aspect of our students'
lives, and many of the ``killer apps'' of the future will lie at the
intersection of the computing and the physical world -- robotics, the internet
of things, personal electronics, etc.  Students prepared to create those
applications must understand how to design, build, and debug systems that
include both code and physical components.
Likewise, programming has become an
integral part of every engineering discipline, so many engineering students can
benefit from hands-on experience building a programming complex systems.
Educators have devised a wide range of capstone courses to provide students
with experience at the boundary of hardware and software~\cite{5937784, 
1509339, 
876968, 
1408742, 
4147435}, but 
providing this experience in the face of growing enrollments is a significant
challenge.

To provide students with project-based, hands-on experience we have developed a 10-week
(one academic quarter) capstone, inter-disciplinary course in which students
design and build all of the hardware and software components of a
remote-controlled quad-rotor aircraft (or ``\q{}'').  The course content
includes embedded system/microcontroller programming, debugging software-controlled mechanical systems, working on interdisciplinary teams, PCB design, basic sensing and control theory, and practical skills in
PCB assembly, soldering, and debugging electro-mechanical systems.  The class
targets senior undergraduate (or graduate) computer science, computer
engineering, and other engineering students.  We have taught the class annually
for the last four years in the computer science and engineering department at the University of California, San Diego.
The class has been
growing in size and most recently had 24 students working in twelve groups.

A particularly challenging aspect of teaching the course is providing detailed,
thorough, and reliable design reviews for student designs.  Even a small error
in the PCB design can lead to failure, and most teams require multiple reviews
before their design is ready to manufacture.  The workload can be crushing for
the instructors.

To remedy this problem, we developed an automatic design checker call
\emph{\eaglint{}} that can verify many aspects of student designs.  The tool
drastically reduces the number of human design reviews required and lets those
reviews focus on less tedious aspects of the design review process.  It is
thorough enough to fully automate grading for some of the preliminary labs in
the course.  We believe \eaglint{} is the first autograder to handle PCB
designs.

apluThe emphasis on building hardware is unusual for a computer science course, but
understanding how to build computer systems is critical for students hoping to craft
next-generation devices that rely on the careful blending of hardware and
software\footnote{Also, as Turing Award winner Alan Kay famously said ``People who are really serious about software should make their own hardware''}.

The course materials are all freely available online\footnote{https://sites.google.com/a/eng.ucsd.edu/quadcopterclass/\\https://github.com/NVSL?q=QuadClass}.  The only exceptions are a
complete ``solution'' to the project and the source code for our automatic
design checker.  We are happy to share these with educators.

Based on surveys, course evaluations, and anecdotal reports, students find the
class to be very challenging, but also exciting, engaging, educational, and
fun.  The course staff reports that it is great fun to teach as well.

This paper describes the course content, the challenges that arise in teaching
the class, our design checking tool, and student reflections on the course.
Most of the descriptions reflect the most recent instance of the class (Spring
2018).  We also detail our plans for the next version which will address many of
these challenges and should allow the class to grow further in size.

\section{Project Overview}
\label{sec:overiew}

\begin{table}
  {\footnotesize
  \begin{tabular}{|p{0.75in}|p{2.25in}|}%
    \hline
H1: Eagle Intro	&	Complete an Eagle tutorial on building libraries, schematic assembly, layout, design checks, and CAM file generation.\ignore{	&	2 days	&	Understand the basics of Eagle operation and the PCB design flow.}\\\hline

H2: PCB Libraries	&	Build several Eagle library parts (of varying complexity) from scratch by reading and interpreting data sheets.\ignore{	&	1 week	&	Understand the content, structure, and best practices for PCB library design.}\\\hline

H3: Schematic \newline creation	&       Create the schematic for their \q{} based on a combination of reference designs, written specifications, and datasheets.  Design an LED-based lighting element.\ignore{	1.5 weeks	&	Understand how to assemble a schematic design from a diverse set of resources that reflect real-world design practice.  Gain experience in manual verification of their work.}\\\hline

H4: PCB Layout	&	Design the shape of their PCB and layout the components to meet mechanical, electrical, and aesthetic constraints.\ignore{	1.5 weeks	&	Understand the principles and concepts of PCB layout and gain experience satisfying diverse design constraints.    Gain experience in manual verification of their work.}\\\hline

H5: PCB Assembly	& Assemble their PCB using solder paste, tweezers, and a reflow oven.\ignore{	1-2 weeks	&	Undertand the principles of PCB assembly and gain experience with soldering and PCB "rework".}\\\hline\hline

S1: Software Intro	&	Setup the Arduino IDE and program their remote control and PID test stand to allow remote control of the motor speed.  Assemble a PID test stand.\ignore{	1 week	&	Learn how to use the Arduino IDE.  Understand how to program all the
basic components of the remote and \q{}.  Understand the components and operation of a brushed motor driver.}\\\hline

S2: Sensing	&	Implement sensor filtering and fusion using software and the IMU's internal filters to provide clean, low-noise measurements of the \q{}'s orientation.\ignore{	1-2 weeks	&	Undrestand how to configure and use a raw hardware device (the IMU).  Understand the basics of sensor filtering and fusion.  Gain experience debugging a software/physical system.}\\\hline

S3: PID	&	Implement a one-channel PID controller to stabilize the PID test stand.   Control the pitch of the test stand using their remote control. \ignore{	1-2 weeks	&	Understand the principles of PID-based control and gain first-hand experience tuning a PID controller.  Gain experience debugging a software/physical system.}\\\hline

S4: Flight Control	&	Implement flight operation functions such as "arming" the \q{} and calibrating of the gimbals and IMU.\ignore{	1 week	&	Understand the details of what's required for safe \q{} flight.  Gain experience in low-level programming.}\\\hline\hline

H6: Flight!	&	Combine the above elements to make their \q{} fly.\ignore{	0.5-3 weeks	&	Gain experience debugging and tuning a complex software/hardware system.  Apply skills acquired in labs S1-S4 in a demanding "real-world" system.}\\\hline
 \end{tabular}}
 \caption{\figtitle{Lab Descriptions} The course comprises ten labs broken into
   three groups.  Many of the hardware and software labs run concurrently to
   fit them all into ten weeks.}
 \label{tab:labs}
\end{table}

\wfigure[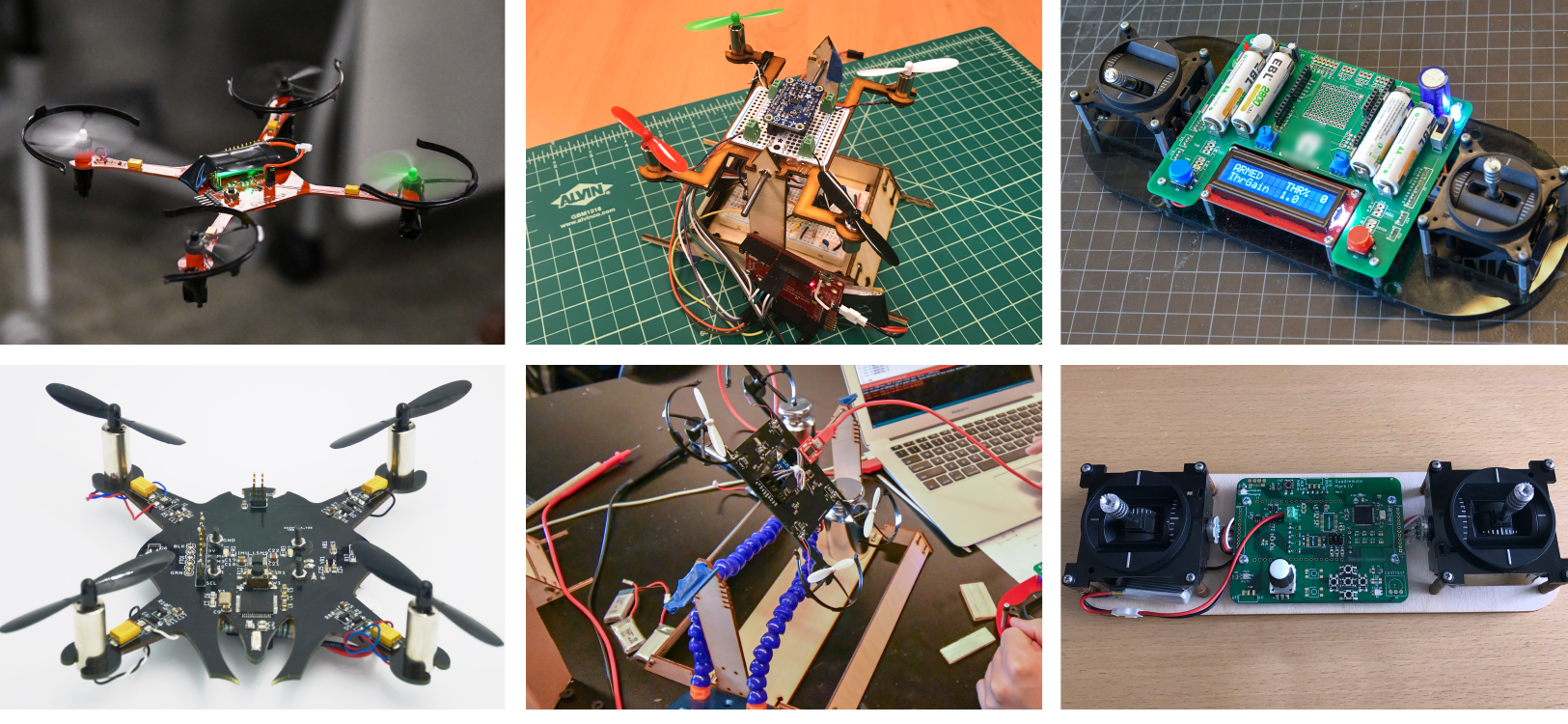,{\figtitle{Photos from the Class} (From left to right)  Two student \qs{} (top and bottom), the old test stand (top) and the new (bottom), and the current, expensive remote control (top) and the cost-optimized version (bottom).},fig:montage]

The class is an intensive, capstone-style course.  It moves very quickly and
students need to stay engaged and work hard to succeed.  With just a very few
exceptions over the last four years, students rise to the course's challenge
with aplomb.

The \qs{} that students design use a PCB to host their electronics and to serve as its
airframe (\reffig{fig:montage}-left).  They measure less than 10~cm on a side and are suitable for flight
indoors over short distances.  The motors are moderately powerful, ``brushed''
electric motors powered by a small lithium-polymer (LiPo) battery, and we use
small, plastic propellers.  The \qs{} are easy to operate safely, and a blow
from the propeller at full speed is painful but not particularly
dangerous.  Students wear eye protection around their flying \qs{}.

The \qs{} use an Arduino-compatible, 16~MHz Atmega128RFA microcontroller~\cite{Atmega128RFA}
that includes an
IEEE~802.15.4~\cite{IEEE802.15.4} radio.  To measure the \q{}'s orientation, we
use an inertial measurement unit (IMU)~\cite{lsm9ds1} that contains a 3-axis
accelerometer and a 3-axis gyroscope.  The purpose-built remote control
(\reffig{fig:montage}-right) uses the same microcontroller, providing a uniform programming environment for the course.

We teach the class in a ``Makerspace'' run by our engineering school to support
hands-on engineering courses.  It provides an array of useful equipment,
including high-quality soldering equipment, a professional reflow soldering
oven, a laser cutter, and an array of hand tools.

\section{Course Content}
\label{sec:overview}

The course breaks building a \q{} into four high-level tasks: designing the
PCB, implementing the flight control software, assembling the PCB, and getting
the \q{} flying.  Below we describe each of these tasks, key challenges they
present for the students and the course staff, and how we approach them in the
class.  Table~\ref{tab:labs} briefly describes the course's ten labs.

\subsection{Designing and Manufacturing PCBs}

The first task introduces students to the key concepts of PCBs design,
the tools and processes used to design and manufacture them, and best practices
for PCB design.

They complete labs \lab{H1} and \lab{H2} on their own.  For the remaining labs,
they work in pairs.  We manufacture the student PCBs through JLPCB~\cite{jlcpcb}.
The boards usually arrive with 5-7 days of
ordering.  We build four-layer PCBs.

\boldparagraph{Key Topics}  This task  addresses the following concepts and skills.

\begin{enumerate}

  \myitem{PCB principles} What PCBs are, how they work, and how they are manufactured.

  \myitem{PCB Design Tools} The abstractions and concepts that PCB design tools provide to describe PCB designs  -- schematics, part libraries, layers, design rules, etc.

  \myitem{Design From Example} How to adapt reference designs to a particular application.

  \myitem{Design From Datasheet} How to interpret device datasheets to integrate a device into a design.

  \myitem{Part Selection} How to select appropriate components for a design from among many, many options.
  
  \myitem{Design Requirements} How to describe, interpret and satisfy electrical, mechanical, and application requirements for a PCB.

  \myitem{PCB Design Best Practices} How to design schematics and PCBs that functional, manufacturable, and comprehensible to other engineers.

  \myitem{PCB Manufacturing Process} How to prepare a design for manufacturing and interpret the design requirements of a particular manufacturer.

  \myitem{The Cost of PCB Errors} The importance of detailed design review and attention to detail in PCB design.
  
\end{enumerate}

The last point represents a significant departure from most other areas of
computer science, where the compile-test-debug cycle is extremely short and
fixing many errors is nearly free.  In the class, an error in a student design
can delay their project by at least two weeks -- an eternity in a ten-week class.  In the real world, PCB design errors can
take months to address.

\boldparagraph{Student Challenges} Students face two types of challenge in this part of the class.

The first is using the tools.  We use the ``standard'' version of Autodesk's
Eagle PCB design package.  It is free for students and relatively simple to
use, but its interface has some rough edges.

Second, some students struggle with the ``correct by inspection'' requirement
that comes with designing PCBs.  For some, this requirement is 
stressful.  Others struggle to pay enough attention to detail while reviewing
their design.

\boldparagraph{Staff Challenges} The main challenge for the
course staff is performing design checks to help ensure the student's PCBs will
work.  The typical \q{} schematic has \~{}190 electrical connections and an error in
any of them can cause the board to fail.  In addition, there are numerous
constraints that the PCB layout must meet in order to work properly.  Checking these is tedious, time-consuming, and the staff can easily overlook subtle problems.  The workload compounds because the students must keep revising their designs
until they are correct.

We address this challenge in three ways.  First, we require the students to
``pay'' for design reviews (see below).  Second, students perform a peer
design review for the schematic and another for the PCB layout.  Third, the
staff only performs careful design checks on the final schematic and layout.  The earlier labs rely mostly on automated design checks (See \refsec{sec:eaglint}).

\boldparagraph{Teaching} The PCB portion of the course moves very quickly and
starts the first day of class.  In quick succession, students learn the basics
of Eagle (\lab{H1} -- 2 days), how to build high-quality PCB libraries
(\lab{H2} -- 1 week), and then design (\lab{H3} -- 1.5 weeks) and layout
(\lab{H4} -- 1.5 weeks) their PCB.

We introduce each topic with a lecture and then describe the lab and demonstrate
the key features of Eagle they will use.  We also describe the electrical
components they will use and discuss how to interpret and use datasheets and
reference designs.

We use an unusual method for grading the hardware labs that reflects the fact
that, in the real world, PCB design errors are expensive.  Students
``pay'' for design reviews (by course staff or \eaglint{}) with points
deducted from their lab grade.  This incentivizes them to find and fix
problems themselves by inspection rather than relying on \eaglint{} or the staff.

The labs are worth 10 points.  We give them 12 points to start and
subtract 0.5 points for each design review they request.  Completing a lab
requires passing both a \eaglint{} review and a human review, so they can score
up to 11 out of 10.  We do not deduct points for anything else on the PCB
design labs -- the lab must be completed correctly for them to move on.  This may seem harsh, but the
alternative is manufacturing a PCB with known flaws, which is pointless.

\subsection{Implementing Flight Control Software}

The flight control software task provides students with
first-hand experience implementing a software system that controls a real-world
device, comprises two communicating components (the remote and the \q{}), and
presents challenging debugging problems.  The two central challenges are 1)
combining the inputs from the gyroscope and the accelerometer to provide
accurate, responsive measurement of the \q{}'s orientation and 2) implementing
and tuning a proportional, integral, derivative (PID)~\cite{pid} controller to control that orientation.

We provide some pre-built equipment for this portion of the course: a
PID test stand (\reffig{fig:montage}~top-middle).  The test stand supports a mockup that
closely matches components of the \q{} they will build.  It has a
microprocessor break-out board (red)~\cite{atmega129rfahookup,atmega128Rfabob}\footnote{Unfortunately, Sparkfun has discontinued this breakout board.  We have a good number of them, but we are migrating away from it (see \refsec{sec:future}).}, an IMU breakout board (blue)~\cite{lsm9ds1bob,lsm9ds1hookup}, and a
laser-cut plywood frame.  It has a pivot that allows the mockup to tip back and
forth, simulating a \q{} that moves on one axis.

We leverage the open-source library support for the IMU~\cite{lsm9ds1lib,sensorlib,ahrs} and
microcontroller~\cite{atmega129rfahookup,atmega128rfalib} provided by SparkFun~\cite{sparkfun} and
Adafruit~\cite{adafruit}.

\boldparagraph{Key Topics} This task addresses the following concepts and skills:

\begin{enumerate}

  \ignore{\myitem{Programming each component} How to use each component of the \q{} and remote (e.g., controlling motor speed, configuring the IMU's built-in filters, reading values from the IMU, sending and receiving data via the radio). }

  \myitem{Attitude sensing} How to measure orientation by combining gyroscope and accelerometer measurements.
  
  \myitem{Sensor Filtering} How to use the IMU's built-in filtering facilities to reduce the burden on software.

  \myitem{Complementary Filtering} How to use a complementary filter to combine sensor inputs to provide more stable, lower-noise measurements.
  
  \myitem{Basic PID control} How PID controllers work and how to implement and tune them in software.
  
  \myitem{Debugging physical systems} How to debug and tune software that controls a physical system.

\end{enumerate}

Items two through four are the subjects of entire courses (which most student have not taken).  We cover the basics and
provide intuition for the underlying theory.

\boldparagraph{Student Challenges} Students struggle with several aspects of
this task.  The first is that the notion of ``correctness'' for both the
filtering and PID code is qualitative rather than quantitative.  We provide
guidance about the how algorithms should behave, but there is not a
crisply-defined ``correct'' answer.

Second, poor performance can stem from many sources: Conceptual
misunderstanding, algorithm implementation errors, arithmetic problems (e.g.,
overflow), misconfiguration of the IMU, and poor parameter settings.   Finding the root cause of a problem can be hard.

\ignore{Another source of poor performance is the breadboard motor drivers.  They can
be unreliable, inconsistent, and under-powered because of poor electrical
connections in the breadboard and the high-currents required by the motors.
These problems are difficult to debug, are out of the scope of the class, and
prevent some groups from achieving good results.}

\boldparagraph{Staff Challenges} The main staff challenge is to help the
students in debugging their code efficiently.  The algorithms are not very
complex, but there are many ways to implement them and getting oriented in each team's code base is impractical.

Instead, when they face a problem, we have students verify that individual
components of the system are working properly starting with the simplest.  For
instance, for trouble with a PID controller, we ask them to verify the
correctness of their raw sensor readings, then their filtering
code, etc. to identify where, exactly, the problem lies.

\ignore{A second challenge is in the wide range of student expertise in control theory.
Some students will have a stronger background than the course staff while
others are seeing it all for the first time.}

\boldparagraph{Teaching} We divide this task  into three labs,
each with an associated lecture.  The first (\lab{S1} -- 1 week) covers the basic
hardware components on the remote and the \q{}.  They assemble code from
example programs to demonstrate that they can read sensor
data from the IMU and control the speed of motors on their test stand using the
remote control.

\lab{S2} (1.5 weeks) and the associated lecture covers sensor filtering.  \ignore{The challenge is
that IMU's two sensors have different strengths: The accelerometer has good
absolute accuracy for the \q{} orientation but picks up vibration
noise from the motors.  The gyroscope is mostly immune to noise, but its
measurements drift over time.} The
students use a complementary filter~\cite{compfilter} combined with the IMU's built-in high- and low-pass filters
to generate accurate, low-noise
orientation measurements. \ignore{The demonstrate their success by moving the test
platform quickly with the motors running and showing their the measurements
meet this criteria.}

In \lab{S3} (1.5 weeks) they implement a PID controller to stabilize and control the pitch of the test stand.
They must implement PID correctly, tune its parameters, and
translate the controller's output to motor power levels.  \ignore{Accomplishing this requires them to confront 
practical challenges of implementing PID.}

\subsection{PCB Assembly}

This task teaches how to assemble a moderately complex PCB.  It
is a delicate and potentially frustrating process.

\boldparagraph{Key Topics} We teach the following skills:

\begin{enumerate}
  \myitem{Reflow soldering} How reflow soldering works and how to achieve good
  results, including how to applying solder paste to the board.

\ignore{  \myitem{Working the solder paste} How to apply solder paste with a plastic
  syringe, how much to use, and how (and how much) to apply it for good results.}

\ignore{  \myitem{Organizing parts for easy assembly} How to organize discrete
  components so assembly is reasonable quick and highly accurate.}

  \myitem{Placing parts} How to place parts precisely with tweezers and how to
  ensure the correct orientation of polarized parts.

  \myitem{PCB rework techniques} How to fix common problems in hand-assembled PCBs.

  \myitem{Hand soldering} How to solder through-hole components by hand.

  \myitem{Flashing the bootloader} How to install the low-level firmware on their microcontroller.

  \myitem{Testing and bring up} How to systematically check that each component
  of their \q{} is functioning properly.
    
\end{enumerate}

\boldparagraph{Student Challenges} Assembling boards is hard for several
reasons.  First, many students have no experience
soldering.  Common problems
include a lack of patience, applying too much solder paste, and 
applying it imprecisely.

Soldering the IMU is especially challenging.  In the latest iteration of the
class, the failure rate was 75\%, and several teams resorted to ``hot wiring''
the IMU breakout board onto their \q{}.  This is a clear area where we need to
provide a better method\footnote{Interestingly, in previous years the success
  rate has much higher.  Several things have changed since then, but the most
  likely culprit seems to be low-quality solder paste.}

\boldparagraph{Staff Challenges} This has emerged as the most labor-intensive section of
the class for the staff, especially since \eaglint{} has taken on many of the design reviews.  The boards come back from the manufacturer in batches, so many groups need intensive, one-on-one
guidance about part placement, etc. at the same time.  This leads to extremely
busy staff, extended lab hours, and a lot of student waiting.

\ignore{It also takes a great deal of time.  During the most recent quarter, the staff
spent 3-5 hours per day, two days a week for 2 weeks working non-stop with
students to assemble their boards.}

\boldparagraph{Teaching} We provide a brief lecture about reflow soldering and
a small group tutorial about applying solder paste and placing parts.

\subsection{Flight}

In the final stage (\lab{H6}) of the course, students combine their PID
controller software with their PCB to create working, flying \q{}.  This
includes extending the flight control software from one-axis (pitch) to three
(pitch, roll, and yaw), loading it onto their PCB, tuning it to achieve stable
flight.  Stable hovering counts as success.

\ignore{The first step is to replicate their PID controller for pitch to handle roll as
well.  These two are identical.  The PID channel for yaw is slightly different,
since it stabilizes angular speed around the \q{}'s up-down axis rather than
attitude.}

\ignore{The test stand we provide is not large enough to accommodate their \q{}, so
they must assemble or build their own.  They show remarkable creativity.  They
can test both pitch and roll on their test stand, but testing yaw is usually
done in the air once pitch and roll work reasonably well.}

\boldparagraph{Key Topics} The key skills covered include the following:

\begin{enumerate}

  \myitem{Integration and testing} How to generalize their PID
  controller to work in a more challenging environment.

  \myitem{Flying a \q{}} How to maneuver a \q{} safely.

\end{enumerate}

\boldparagraph{Student Challenges} Student success during for stage varies widely.  Some students get it working quickly, but others struggle, and many do not succeed.  In
some cases, failure has a clear cause (e.g., their PID from \lab{S2} did not
work well).  In other cases, the cause is less clear.  It would be useful to
perform interviews with teams to understand what went wrong.

\ignore{One common source of frustration are the differences between the breadboard
motor driver circuit they use in the earlier labs.  These difference usually
necessitate significant re-tuning of their software.}

\boldparagraph{Staff Challenges} The challenges here are similar to those for
the flight controller portion of the class.  The staff mostly provide debugging
and moral support to frustrated students.

\boldparagraph{Teaching} There is no additional material to present for this
portion of the class.  Students start this lab when they have finished \lab{H5}
and \lab{S4} and work on it until the end of the quarter.  In the best case,
they have three weeks.  Some have just a few days.

\section{Logistics}
\label{sec:logistics}

Running the course successfully has required us to carefully manage class size
and the course schedule.  We have also worked to craft a grading scheme to
reward student effort and account for the challenging nature of the course.

Please see the appendix of~\cite{quadcopter-arxiv} for a more detailed
discussion of course logistics.

\subsection{Admissions}

Students must apply to take the course.  We advertise to graduate and
undergraduates across the school of engineering, drawing students mostly from
computer science, electrical engineering, mechanical and aerospace
engineering, and math.

We accept students based on some demonstration of success in projects-based coursework and GPA.  We also prioritize students graduating before the next iteration
of the class.  The last time we taught the class (Spring 2018), we had 119
applicants, accepted 41, and 24 ended up attending the class: Twelve computer
science, computer engineering, or robotics students, nine mechanical or
aerospace engineering, and three math.

\subsection{Schedule}

\cfigure[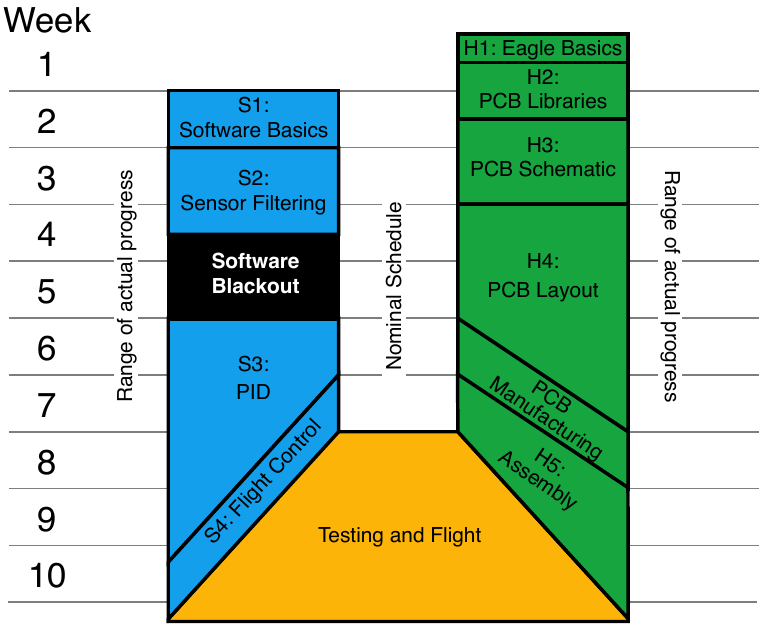,{\figtitle{Course Schedule} The schedule for the
    class moves quickly but accommodates delays.  The nominal schedule (down
    the center) leaves over three weeks for students to get their \qs{} flying.
    In practice, the timeline varies widely between groups (the
    edges).},fig:schedule]
      
The course schedule is very tight, since our university runs on 10-week
quarters.  We meet twice a week with a 1.5-hour lecture section followed
immediately by 1.5 hours of lab time.  The boundary between the lecture and lab is blurry.

We run the hardware and software labs in parallel.  We
lecture about and then start a new lab roughly every day for the first three
weeks of the quarter.  From that point on, groups progress at different rates.

\reffig{fig:schedule} illustrates the nominal schedule (down the center
of the figure) and the slippage that invariably occurs (the outer edges).  The
largest delays are in finalizing the PCB layout, assembly, and completing the
PID controller and flight software.

In earlier versions of the course, we noticed that students (especially
computer science majors) would devote too much time to \lab{S2} rather than
finalizing their PCB layouts.  To prevent this, we added a ``software
blackout'' when they must focus only on their PCB layouts and the 
course staff refuses to answer questions about the software labs.

We do not enforce deadlines on any of the labs after \lab{H3}.

\subsection{Grading}

Aside from the ``pay-for-review'' grading mechanism described above, grading in
the course is a combination of participation (showing and doing work),
and ``checking off'' the completion of the labs.  Our experience is that
students, almost without exception, have worked hard and learned a great deal,
even if their \q{} does not fly in the end.  Generally speaking, working hard
in the class will earn a ``B'', and any semblance of flight earns
an ``A''.  Stable flight rates an ``A+''.

\subsection{Staffing}

The course staffs usually consists of the instructor and one TA.  When we have
taught the class, the instructor has been deeply involved in designing and
running the labs.  The staff needs to be familiar with all of the course
content and have practical experience with designing, assembling, and debugging
PCBs.  Experience with PID control is also very useful, although that (and the
other software components) are easier to pick up ``on the fly,'' if the staff
has general experience with Arduino programming, since errors are less costly
and debugging is easier.

The best preparation for teaching the class is doing the project
start-to-finish.

\subsection{Equipment}

The course requires hand-soldering, reflow soldering, and the facilities to
build a test stand.

A good soldering iron is necessary, but not terribly expensive.  We use
soldering irons similar to the Hakko FX-888D (\$110).

For reflow, we use a high-end reflow oven, but this is not necessary.  The
first three iterations of the course used a ``heat gun'' (i.e., paint stripper)
instead of the reflow oven and the results were very good.  There is also a
rich DIY reflow soldering community online.

We use the laser cutter to build the test stands, but in earlier iterations we
3D-printed the test stands or let the students devise their own testing rig.
They showed remarkable ingenuity with materials including rubber bands,
cardboard, and popsicle sticks.

\section{Automated Design Checks}
\label{sec:eaglint}

Checking PCB schematics and layouts is a time consuming, error-prone, and
critical to maximizing student success in the class.  In the earlier versions
of the course, the course staff would quickly become overwhelmed with design
reviews.  This led to uncaught errors and exhaustion.  The effort required
effectively limited the number of students we could accommodate in the class.

To reduce this burden, we developed a custom, web-based course management tool
called \eaglint{} that performs automated checks on student PCB schematics
and layouts and some other useful classroom functions (e.g., tracking student progress and submitting labs).

\eaglint{} is, we believe, the first autograder that checks specific design
requirements for PCB designs.  All PCB design tools provide a suite of design
rule checks (DRC) and some PCB manufacturing houses provide automated tests for
manufacturability~\cite{freedfm}, but \eaglint{} provides an extensible,
programmable mechanism for checking the higher-level correctness of designs.

\subsection{Checking Designs}

\eaglint{} has two modes.  The \emph{quick check} mode generates warnings that
flag common violations of good PCB/schematic design style, similar to the
checks \texttt{lint} performs on source code.  These style checks could be
applied to any PCB design.  Failure to fix a warning will not, in itself, negatively impact
the function of the PCB.  Students can run quick checks as frequently as they want.

\eaglint{}'s \emph{full check} mode identifies errors that are likely to cause their PCB
to not work correctly.  Most of these checks only apply to \qs{} designed in
this class: They are extremely specific to the reference designs, datasheets,
and specifications we provided.  The students ``pay'' for each full check
with 0.5 points off their score for the current lab.  This prevents them from
relying on \eaglint{} -- if they ever design a PCB on their own, they will not have
the benefit of \eaglint{}'s detailed checks to catch their errors.

\eaglint{} lets students explain why they feel a particular error or warning is
unjustified.  Once they have fixed or explained all their errors and warnings,
they submit the design for human review (which costs another 0.5 points).  The
course staff looks at their explanations and can approve or reject them and
provide written feedback.  If errors remain, the students fix them and
resubmit.  If there are no errors, they move on to the next lab.

\eaglint{} can perform a nearly complete check of student schematics (i.e.,
\eaglint{} passes, it will work).  Checking PCB layouts is more difficult
because many of the requirements are geometric or spatial.

\ignore{\eaglint{} knows which lab each team is working and applies an appropriate set
of checks for their current lab.}

\ignore{\subsection{Other Functions}

\eaglint{} has several other useful features:
\begin{itemize}

  \item We use GitHub Classroom~\cite{githubclassroom} to distribute materials and submit designs and
    source code, and \eaglint{} fetches the designs from GitHub.  This is
    convenient and ensures that students have checked in their working designs.

  \item It automatically pairs groups for peer design reviews.

  \item It tracks student progress through the labs.

  \item It generates combined bills of material (BOMs) for multiple designs to
    make ordering parts easier.

\end{itemize}}

\subsection{Implementation}

\ignore{It is easy to add new checks to \eaglint{} using its declarative programming
interface.  A full description of \eaglint{}'s interface for specifying checks
and mechanism for verifying them is beyond the scope of this paper.}

\eaglint{} runs in the cloud on Google's AppEngine~\cite{appengine}.  It is
written in Python and relies heavily on the Swoop library~\cite{swoop} for
reading and manipulating Eagle PCB design files.  We are happy to share the
source code with educators.

\section{Student Feedback}
\label{sec:results}

Based on anonymous course evaluations administered at the end of the class,
students enjoy the class and find it valuable.  Among 12 respondents from the
most recent class of 24 students, 100\% of students rated their enjoyment of the class a 4 (``some'') or 5 (``a lot'') out of 5.  93\% of the students answered
``yes'' or ``probably'' (4 or 5 out of 5) to whether they could build a PCB on their own, all of
them felt more confident soldering, and 91\% felt more confident in debugging
electronics.  70\% of students thought the pace of the course was ``just
right'' or ``could have been a bit faster'', and the remainder felt it ``could
have been a bit slower.''

\section{Future Improvements}
\label{sec:future}

Planning is already underway for the next iteration of the course, and we are
making some significant changes and smaller adjustments to improve
the student success rates and reduce (or least more evenly distribute) work for
the staff.  Below we detail the planned changes for each of course's four main task.

\subsection{Designing and Manufacturing PCBs}

We are continuing to refine \eaglint{} to catch more common (and uncommon)
errors in student designs and we are also adding features that will make it
easier for less expert course staff to perform design reviews.  We are working on
ways to make the peer-design reviews more useful and effective.

\subsection{Implementing Flight Control Software}

The breakout board-based PID testing setup leads to several problems:
Building it takes time and the electrical connections can be unreliable which leads
to problems in the software labs.

We are addressing all of these problems by replacing the breakout-board based
design with a newly-designed PCB that integrates all the components of their
final design.  This will attach to laser-cut arms to hold motors.  The result
will essentially be a \q{} that fits on a newly-designed, larger test stand
(\reffig{fig:montage}-bottom-middle) that will also accommodate their final
\qs{}.  The new board will use the same components and circuits as their final
designs, so the work they do in the software labs should transfer better to
their \qs{}. \ignore{We are also considering an easier-to-use IMU~\cite{betterimu2}.}

\subsection{PCB Assembly}

The biggest problems we face in PCB assembly are the low success rate for
soldering IMUs, the huge spike in staff workload that occurs during the
assembly process, and the amount of time it takes students to assemble their
\qs{}.  We plan several changes that will address these challenges.

The first is that we have been experimenting with better soldering techniques
for the IMU and considering different IMUs that come in easier-to-solder
packages~\cite{betterimu1}.

The second is that we will have students assemble the PCB for their remote
control.  The remote control assembly lab will run early
in the course, and we will divide the class into 2-3 cohorts that will do the lab
at staggered times, to make teaching and supporting them more manageable.  We
will provide ``loaner'' remotes to ensure that groups can make progress on
other labs if assembly goes poorly.

Finally, we will use stencils to apply solder paste to the PCBs
instead of plastic syringes.  It is faster, more precise, yields better-looking
PCBs, and is how ``real'' PCBs are assembled.

\subsection{Flight}

No significant changes are planned, but the changes outlined above should leave
students with more time at the end of the class to get their \qs{} flying.

\subsection{Cost}
\label{sec:cost}

\cfigure[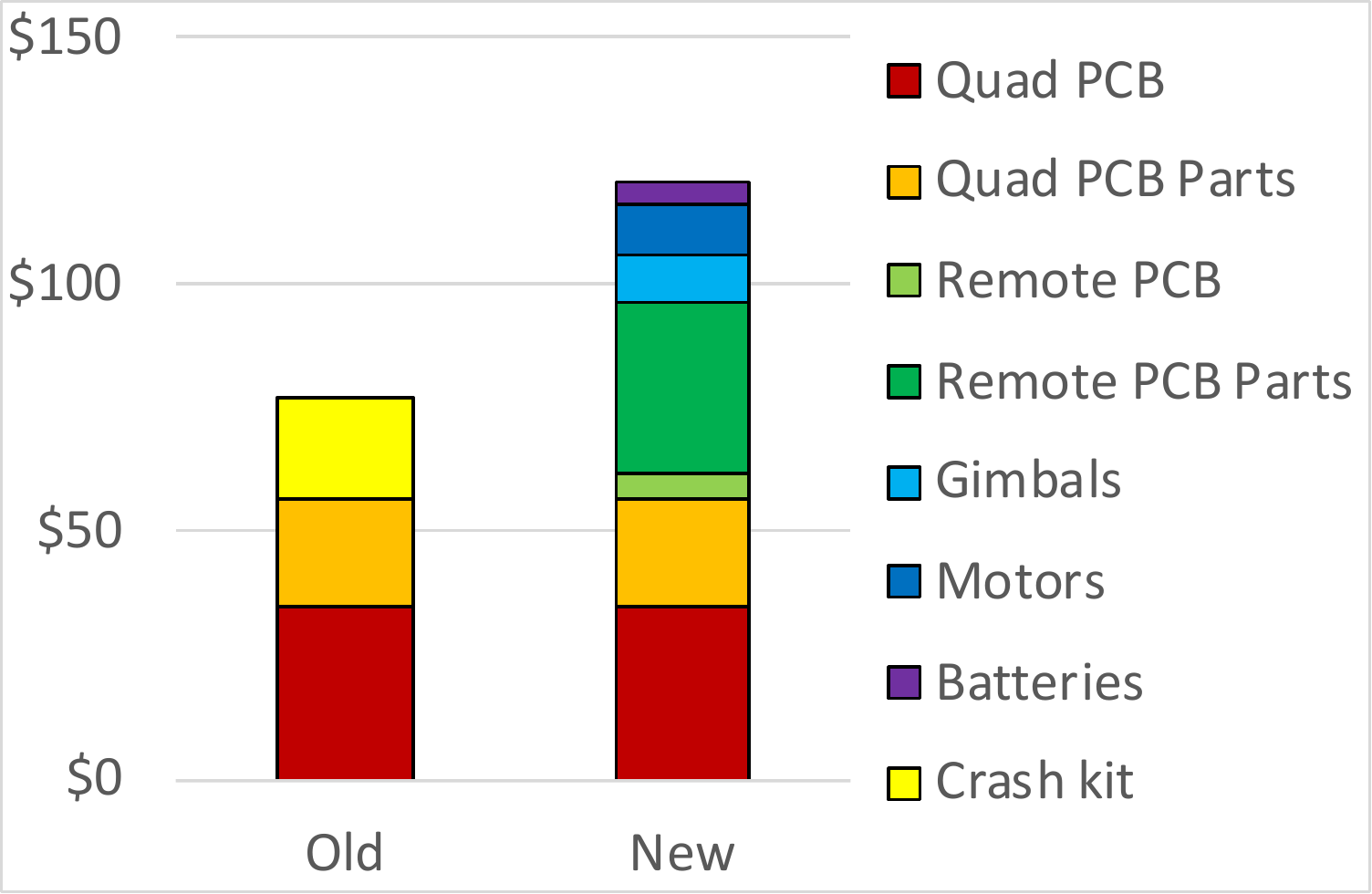,{\figtitle{Per-student Costs} The
    per-student cost for the most recent iteration (left) did not include the
    remote.  Our planned changes will raise costs, but students will
    have everything they need to program their \q{} after the class.},fig:costs]
  
A frequent complaint about the course is that the students cannot continue
working on their \qs{} after the course ends since they cannot keep the remote
controls.  The reason for this is that the remote controls (\reffig{fig:montage}-top-right) we have been using
are expensive -- \~\$150.  The \q{} costs about \~\$35 per board plus \$40 in
for the components (including a ``crash kit''~\cite{crashkit} that provides batteries, motors,
etc. as spare parts for a small, commercial \q{}), so charging a lab fee large
enough to provide a full set of hardware for each student to keep is not
feasible.

We have set a goal of reducing costs enough to let each student keep a
remote, a \q{}, and all the other necessary equipment for
\~\$120.  \reffig{fig:costs} shows the
breakdown in per-student cost for the last iteration of the class and the
projections for the new version.  Note that the per-student cost goes up, but in the
new version, they can keep the remote.

The new remote (\reffig{fig:montage}-bottom-right) dispenses with
useful-but-expensive on-board LCD for displaying status information (\$25).  We
also added an integrated LiPo battery charger, so they can charge their own
batteries.  The remote uses the same kind of battery as the \q{}.  Further
savings come from buying electrical components~\cite{digikey},
motors~\cite{motors}, and batteries~\cite{batteries} in larger quantities
(i.e., enough for 2-3 years of the course).  Ordering in small quantities costs \$10-\$20 per student.

\ignore{\subsection{Admissions}

To date, the amount of equipment available for the class has placed a hard
limit on enrollment, since a key piece of hardware has been discontinued.  The
changes above remove that limitation.  In future classes, we plan to admit more
freely.  We feel that we scale the class to 40-50 students.}

\section{Conclusion}
\label{sec:conclude}

Many of the ``killer apps'' of the near future will lie at the intersection
hardware, software, sensing, robotics, and/or wireless communications.  Building a \q{}
gives students hands-on training in all of these aspects of system design
simultaneously.

Building a \q{} from scratch is 10 weeks is challenging, and fitting all of the
necessary material into a single academic quarter has required careful planning
and refinement over several iterations of the course.  Students almost
universally enjoy the class and report they learn many useful skills over the
course of the quarter.

There is still much room for improvement.  We are
planning changes that should increase student success rate and allow us to
scale the class to meet the growing demand for project-based courses from
students.

\section*{Acknowledgments}

We would like to thank the reviewers for their constructive comments and Leo
Porter for his comments and guidance.  We would also like to thank Jorge Garza
for TAing the course several times and debugging many, many problems.

\bibliographystyle{plain}
\bibliography{local,libpaper/common}

\newpage
\appendix

\section{Detailed Project Notes}

There are a bunch of logistical issues and smaller-scale design decisions we
have made as we teach the course.  Below, we have provided our observations and
experience on the some of these issues.  The information is up-to-date as of
this writing (January 2019).

For details about the course content, please see the course web page\footnote{https://sites.google.com/a/eng.ucsd.edu/quadcopterclass/}.  It includes links to
the github repos for the current course content, including reference designs,
etc.  It also includes extensive discussion of the quadcopter design itself.
We have not replicated any of that information here.

Three of the repos are of particular interest:

\begin{itemize}
\item https://github.com/NVSL/QuadClass-Resources -- The CAM job files, design rule files, and other useful tools and configuration files.  The lab write ups and course slides, too.  Start with README.md.
\item https://github.com/NVSL/QuadClass-Remote -- The design files for the remote control.
\item https://github.com/NVSL/QuadClass-Quadcopter-Solution -- The design for both a complete quadcopter and the controller board we use on the test stand.  It is not publicly visible, since it is the ``key'' for the course project.  We are willing to share it with instructors.
\end{itemize}

We also maintain a spreadsheet with sources for various pieces of equipment we
use in teaching the class (aside from the electronic components for the PCBs,
which are included in the design files).  It might save you some googling\footnote{https://goo.gl/4pRjeG}.

If you have any questions, please contact swanson@cs.ucsd.edu.

\subsection{4-layer vs. 2-layer}

The students build 4-layer boards.  It would probably be possible to build the quadcopters in two layers, but 4-layers offer several advantages:

\begin{enumerate}
\item It allows us to follow common industry practices of having dedicated power and ground planes.
\item It makes layout and routing easier.
\item It lets us to easily build high-current power delivery to the motors.
\end{enumerate}

In the early versions of the course, we had a signficant problems with noise
from the motors and motor drivers interfering with the power supply to the IMU.
The current design solves this problem by providing large, separate power
planes for the motors.  This would probably not be possible with 2 layers.  Four layer
boards are somewhat more expensive that two-layer boards, but it is a small
fraction of the overal costs for the course.

\subsection{Board manufacturing logistics and cost}

We use JLCPCB\footnote{https://jlcpcb.com} to manufacture the boards.  They are fast and reasonably cheap.
We have not had any quality issues.  They are based in China which raises the
possibility of boards being stuck in customs for an indeterminant period.  This
has happened only once out of perhaps 50 orders we have placed.  We order the
1.6mm boards with lead-free ENIG finish. 
costs more, but seems to help with soldering the IMU.  They also provide cheap
(\$6) solder stencils.  They also build boards in multiple colors, which students enjoy.

There are wide range of other vendors available.  We have used 4PCB (the only
fully US-based option we are aware of), but they are more expensive, no faster,
and don't sell stencils.

\subsection{Brushed motors vs. Brushless}
Most quadcopters use brushless motors that are more durable and more powerful
than the brushed motors that our quadcopters (and most ``micro'' quadcopters)
use.  There are several practical reasons for this choice:

\begin{enumerate}
\item Brushless motors require more complex drivers.  In practice, each brushless motor requires a separate electronic speed conrol (ESC) module or circuit.  These are expensive since each includes a dedicated microcontroller.  Brushed motor are very simple to drive.
\item Brushed motors are adequate for the level of flight capabilities our quadcopters attain.
\item Brushless motors are powerful enough to be dangerous.
\end{enumerate}

During one section of the course an ambitious group persuaded me to let them
use ESCs and more powerful motors.  One of them sustained a significant injury
to one finger.  The motors and props we use are not, in our experience,
powerful enough to cause significant injury.

\subsection{Motors}
We use 8.5x20 (8.5mm in diameter, 20mm long) brushless motors.  They are
designed for high-performance mini-quadcopters that are lighter and more agile
than ours.  They provide more power than we need, but that avoids the need for
us to worry significantly about weight.  They are avialable in bulk from
makerfire.com\footnote{https://makerfire.en.alibaba.com/product/60769263399-802423903/Smart\_4pcs\_8\_5\_20\_motor\_16000kv\_1\_0mm\_shaft\_JST1\_25\_connector\_80mm\_cable\_.html}, but they are not listed on their web site.  

The only slightly unusual thing about the motors we use is that they have plugs
on the end of the wires rather than bare leads.  Students have surprising
difficult soldering wires to their boards, and this fixes that.  The downside
is that you need a mating connector on the PCB.  One annoying problem is that
the motors are listed as having a JST 1.25mm connector (JST is a connector
manufacturer).  The plugs (and the mating connectors) are actually built by
Molex (a JST competitor).  ``JST'' seems to be a hobbyist catch-all for ``small
plastic connector''.

\subsection{The IMU}
There are a wide range of IMUs available.  We chose the LSM9DS1\footnote{https://www.st.com/en/mems-and-sensors/lsm9ds1.html} because
a solid Arduino library and development board is availble from Adafruit.com\footnote{https://www.adafruit.com/product/3387}.
It is also convenient that it includes both a gyroscope and accelerometer.
This eliminates the package count for the quadcopter by one.
 
We are considering two other IMU solutions.  The first is to use separate
gyroscope (FXAS21002\footnote{https://www.nxp.com/products/sensors/motion-sensors/gyroscopes/3-axis-digital-gyroscope:FXAS21002C}) and accelerameter (FXOS8700\footnote{https://www.nxp.com/products/sensors/motion-sensors/6-axis-sensors/digital-motion-sensor-3d-accelerometer-2g-4g-8g-plus-3d-magnetometer:FXOS8700CQ}) ICs.  Adafruit sells a break
out board\footnote{https://learn.adafruit.com/nxp-precision-9dof-breakout} with this
configuration.  This combination has gotten strong reviews and the parts both
have exposed leads which would make soldering easier.  However, it would
increase component count.

The second option is the BNO055, which has an integrated microcontroller that
performs attitude calculations\footnote{https://www.adafruit.com/product/2472}.
There's an Adafruit board\footnote{https://www.adafruit.com/product/2472}.  It
would simplify/eliminate the need for the students to worry about sensor
fusion, but requires additional components on the PCB (e.g., a dedicated
crystal).

\subsection{The Microcontroller}

The key characteristics of the microcontroller we are 1) it is Arduino
compatible and 2) it has an integrated radio.  There are, of course, many
alternative microcontrollers available, but most would require a separet radio
IC, increasing component count and cost.

The most attractive alternative is the Bluetooth LE-equipped Nordic
nRF52832\footnote{https://www.nordicsemi.com/Products/Low-power-short-range-wireless/nRF52832}. There's an Adafruit
board\footnote{https://www.adafruit.com/product/3406} and one from Sparkfun\footnote{https://www.sparkfun.com/products/13990}.  It has plenty of pins,
runs at 64MHz and has hardware floating support, and there's Arduino support.
The biggest advantage, though is BLE: Students could control their quadcopter
from their phones, eliminating the need for a remote control, drastically
reducing the cost of the course.

\subsection{The Remote Control}

There are many low-cost remote controls available for quadcopters.  We built a
custom remote controller around the same microcontroler we use on the
quadcopter for several reasons:

\begin{enumerate}
\item Commercial remotes use a propriatary protocol.  There are open-source projects to reverse engineer the protocol, but even then, it would require an additional (hard to acquire) chip on the quadcopter.
\item Building our own remote allows the students complete control over the remote and quadcopter firmware.
\item Using the same chip provides a uniform development environment across both the remote and quadcopter.
\item Assembling the remote PCB provides a chance for the students to practice PCB assembly.
\item Our new remote design includes several features that reduce the overall cost of the course materials (see below).
\item Our remote uses the same LiPo battery as the quadcopters.
\item The remote control board and schematic are a good reference design for the microcontroller and antenna driver that the students use to guide their quadcopter design.
\end{enumerate}

Our remote integrates a LiPo battery charger that draws power from a 5V FTDI
USB/serial converter.  This lets the students charge their own batteries, which
simplifies class logistics (we don't have to provide access to a charger and they can charge at home).

The remote support a two configurations.  The ``basic'' configuration connects
to a computer via a separate USB/Serial converter that is used to program the
remote.  The FTDI USB/Serial converters we use are expensive (~\$15), so
students don't keep them.  This keeps costs down but means students can't
reprogram their remote or easily charge their batteries after class ends.

The basic configuration also minimizes part count by making some unusual design
decisions.  For example it reduces the number LED packages by using dual-LEDs.
It also doesn't have a power switch -- students just unplug the battery.

The ``advanced'' configuration is more expensive, but has more features.  It
adds an builtin FTDI USB/Serial converter and a USB connection.  The FTDI
connects to the on-board microcontroller \emph{and} to an FTDI break out
header.  This mean that by flipping a switch, the students can program the
remote control or use the remote as an USB/Serial converter to program the
quadcopter.  It can also host a fancy LCD
display\footnote{https://www.sparkfun.com/products/14072}.

We let interested students upgrade their remotes to the advanced configuration
during the course.  It gives them a chance to practice more advanced soldering
skills.

The remote frame is built to be laser cut from 1/8th-inch plywood\footnote{https://www.amazon.com/gp/product/B016H589HC/ref=oh\_makeaui\_search\_asin\_title?ie=UTF8\&psc=1}.  It
accomodates the PCB and two gimbals (joy sticks).  We attach the PCB and
gimbals using long machine screws.  We laser cut plywood ``donuts'' to use as
standoffs, as metal stand offs are surprisingly expensive.

The remote uses gimbals available as spare parts for the larger, high-end
remotes for RC aircraft~\footnote{https://hobbyking.com/en\_us/turnigy-9x-9ch-transmitter-replacement-throttle-rudder-gimbal-1pc.html and https://hobbyking.com/en\_us/turnigy-9x-9ch-transmitter-replacement-elevator-aileron-gimbal-1pc.html?\_\_\_store=en\_us}.  They are readily available, reasonably cheap (\$5
each), and built for control quadcopters.  There are micro joysticks available
for Digikey and other sources.  They are smaller, harder to control, and not
substantially cheaper.  The PCB also has connections for higher-end gimbals
from FrSky\footnote{https://hobbyking.com/en\_us/frsky-replacement-gimbal-for-taranis-transmitter.html?\_\_\_store=en\_us}.

\subsection{PCB Design Tools}
One of the most frequent questions we get from students is about using higher-end PCB design tools rather than Eagle.  Eagle is certaintly not the best PCB design tool available, but it has several advantages:

\begin{enumerate}
\item Students can get free access to it.
\item Its learning curve is managable in a course.
\item There is broad support for it online.
\item Its file format is open and easy to parse, which is critical for our PCB auto-grader tool.
\item It is adequate for our needs.
\end{enumerate}

In addition, Eagle was recently acquired by Autodesk, and has benefited from
significant upgrades in last few versions.

\subsection{Test Stand}

We have redesigned the test stand.  It will use a square flight
control board that is essentially identical to what the students will build.
It will attach to a lasecut frame with arms to hold the motors, creating a
simple clunky-looking quadcopter.

This will mount to a pivot that lets the quadcopter rotate on one axis.  The
control board has FTDI headers on the bottom, so the students can connect via
serial without interfering too much with the quadcopter's movement.

The control board is in the private ``solution'' repo listed above.  The
test stand is on github\footnote{https://github.com/NVSL/QuadClass-Test-Stand}.  It's
authored in SolidPython\footnote{https://github.com/SolidCode/SolidPython}.
The main pieces friction fit together so it's easy to put up and take down.

\end{document}